\documentclass[a4paper,fleqn]{cas-sc}

\usepackage{amsmath,amssymb,amsfonts}
\usepackage{graphicx}
\usepackage{caption}
\usepackage{subcaption}
\usepackage[export]{adjustbox}
\usepackage{threeparttable}
\usepackage{booktabs}
\usepackage{multirow}
\usepackage{lipsum}
\usepackage{makecell}
\usepackage{lineno}
\usepackage[numbers]{natbib}
\usepackage{hyperref}

\def\tsc#1{\csdef{#1}{\textsc{\lowercase{#1}}\xspace}}
\tsc{WGM}
\tsc{QE}

\begin{document}
\let\WriteBookmarks\relax
\def\floatpagepagefraction{1}
\def\textpagefraction{.001}
\let\printorcid\relax 

\shorttitle{Enhancing ASD Early Detection with PCD Protocol and Attention-enhanced GCN-xLSTM Framework}   

\shortauthors{Xiang Li et al.}

\title[mode = title]{Enhancing Autism Spectrum Disorder Early Detection with the Parent-Child Dyads Block-Play Protocol and an Attention-enhanced GCN-xLSTM Hybrid Deep Learning Framework}

\author[1,2]{Xiang Li}
\fnmark[1]
\author[3]{Lizhou Fan}
\fnmark[1]

\fntext[1]{The two authors contribute equally to this work.}
\author[1,2]{Hanbo Wu}
\author[4]{Kunping Chen}
\author[4]{Xiaoxiao Yu}
\author[4]{Chao Che}
\author[4]{Zhifeng Cai}
\author[4]{Xiuhong Niu}
\author[4]{Aihua Cao}
\author[1,2]{Xin Ma}
\ead{maxin@sdu.edu.cn}
\cormark[1] 

\address[1]{School of Control Science and Engineering, Shandong University, Jinan 250061, Shandong, China}
\address[2]{Engineering Research Center of Intelligent Unmanned System, Shandong University, Jinan 250061, Shandong, China}
\address[3]{School of Information, University of Michigan, Ann Arbor 48109, USA}
\address[4]{Department of Pediatrics, Qilu Hospital of Shandong University, Jinan 250061, Shandong, China}

\cortext[1]{Corresponding author}

\begin{abstract}
Autism Spectrum Disorder (ASD) is a rapidly growing neurodevelopmental disorder. Performing a timely intervention is crucial for the growth of young children with ASD, but traditional clinical screening methods lack objectivity. This study introduces an innovative approach to early detection of ASD. The contributions are threefold. First, this work proposes a novel Parent-Child Dyads Block-Play (PCB) protocol, grounded in kinesiological and neuroscientific research, to identify behavioral patterns distinguishing ASD from typically developing (TD) toddlers. Second, we have compiled a substantial video dataset, featuring 40 ASD and 89 TD toddlers engaged in block play with parents. This dataset exceeds previous efforts on both the scale of participants and the length of individual sessions. Third, our approach to action analysis in videos employs a hybrid deep learning framework, integrating a two-stream graph convolution network with attention-enhanced xLSTM (2sGCN-AxLSTM). This framework is adept at capturing dynamic interactions between toddlers and parents by extracting spatial features correlated with upper body and head movements and focusing on global contextual information of action sequences over time. By learning these global features with spatio-temporal correlations, our 2sGCN-AxLSTM effectively analyzes dynamic human behavior patterns and demonstrates an unprecedented accuracy of 89.6\% in early detection of ASD. 
Our approach shows strong potential for enhancing early ASD diagnosis by accurately analyzing parent-child interactions, providing a critical tool to support timely and informed clinical decision-making.
\end{abstract}



\begin{keywords}
Autism spectrum disorder \sep 
biomarker \sep 
behavior recognition \sep 
skeleton data \sep 
GCN \sep 
xLSTM
\end{keywords}

\maketitle




\section{Introduction}

Autism Spectrum Disorder (ASD) is a multifaceted neurodevelopmental condition marked by social communication challenges and behavioral abnormalities. The prevalence of ASD in the United States has escalated sharply, from 1 in 150 individuals in 2002 to 1 in 44 in 2021 \cite{baio2018prevalence}. In China, more than 10 million people live with ASD, with annual diagnoses increasing by approximately 200,000 \cite{yuan2022early}. This growing incidence underscores the emergence of ASD as a significant global public health issue, with profound impacts on affected individuals, family dynamics, and societal resources in both the economic and healthcare sectors \cite{wallace2012global}.

Children diagnosed with ASD demonstrate distinct atypical social behaviors. These behaviors include a diminished engagement with social stimuli \cite{dawson1998children} and an absence of preferential attention to social cues \cite{dawson2004early}, observable as early as six months of age \cite{werner2000brief}. Although such behavioral patterns are not exclusively associated with ASD, they are posited as potential biomarkers for ASD and other developmental anomalies \cite{constantino2017infant}. Moreover, the dynamics within the parent-child interactive relationship, specifically parental behavior, substantially influence the developmental trajectory of children with ASD and are integral to various therapeutic interventions \cite{crowell2019parenting}. In addition, kinesiological and neuroscientific evidence demonstrates that early detection can significantly increase awareness of pathologic behaviors of ASD, and early intervention could improve ASD toddlers’ communication skills and language ability \cite{dunlap2019autism}. Moreover, the current diagnostic process, which relies on subjective assessment tools such as scales and questionnaires, is not only time-intensive but also requires highly trained clinicians to interpret behavioral histories and parental accounts. A competent clinician's evaluation of a case with ADOS \cite{dilavore1995pre} and ADI-R \cite{lord1994autism} typically exceeds ten hours \cite{falkmer2013diagnostic}. In addition, these ‘gold standard’ assessments are prone to biases: subjective judgments can affect interactions during assessments, and interviews based on tools such as ADI-R are highly dependent on the precision of a parent's memory and comprehension \cite{randall2018diagnostic}.

Computer vision (CV) technologies have been increasingly employed to analyze and recognize human behaviors, offering a more objective and efficient means of detecting ASD. Researchers have developed various protocols, such as Response to Name \cite{liu2020early} (RTN), Expressing Needs with Index Finger Pointing \cite{wang2019ENIFP} (ENIFP), and Robot-Assisted protocol \cite{marinoiu20183d} (RAP), to analyze critical behaviors like head, finger, and facial movements. These protocols are instrumental in evaluating joint attention and the quality of social communication, augmented by CV capabilities. In addition, trials have used biomarkers such as eye tracking \cite{murias2018validation}, head movements \cite{martin2018objective}, and motor movement \cite{negin2021vision} to identify ASD characteristics. However, protocols that focus on single biomarkers may not fully capture the complexities of social interaction and cognitive behavior, particularly those involving a sequence of observable complex actions and decisions that unfold during social exchanges and cognitive tasks. Traditional scales such as ADI-R, ASIEP-3 \cite{cordeiro2020evaluating}, or STAT \cite{stone2000brief}, though commonly used to assess ASD symptoms, require lengthy direct observations by clinicians.

To address the limitations of current ASD screening methods, we have developed the Parent-Child Dyads Block-Play (PCB) protocol. Inspired by ADOS-2 indicators, the PCB protocol is designed to assess social interaction, intention, and engagement, focusing on mutual attention and characteristic hand movements in the natural interaction between children and their parents. Applying the PCB protocol to video action analysis, we then propose the two-stream GCN-AxLSTM network, a deep learning framework specifically to aid in the diagnosis of ASD in toddlers to reduce the impact of subjective diagnosis by physicians. This framework has been evaluated using the specially collected PCB4ASD-ED dataset, which includes 129 participant interactions. Consequently, this paper presents a cutting-edge method for helping diagnose ASD in toddlers through video action analysis, emphasizing reciprocal social interaction. The primary contributions of this paper are as follows.

\begin{itemize}
    \item  \textbf{Parent-Child Dyads Block-Play (PCB) protocol.} PCB is an innovative protocol that aims to capture specific ASD-related behaviors in toddlers, providing standardized guidelines for consistent future assessments. 
    \item  \textbf{PCB4ASD-ED video dataset.} PCB4ASD-ED is a comprehensive annotated dataset that is more extensive in both the number of participants and the duration of individual sessions than previous datasets. This dataset documents interactive behaviors of toddlers, which serves as a crucial resource for fine-grained behavioral analysis in early ASD screening. 
    \item \textbf{2sGCN-AxLSTM framework, a hybrid deep learning framework for skeleton-based behavioral analysis in videos.} This framework combines two-stream graph convolution with attention-enhanced xLSTM, to extract spatio-temporal features from long-duration behaviors, enhancing the screening process for ASD in toddlers using the PCB4ASD-ED dataset.
    
\end{itemize}

\section{Related Work}
In recent years, there has been a surge in research leveraging CV techniques for the detection of ASD, with various machine learning and deep learning algorithms developed to discern behaviors from established behavioral markers 
\cite{chen2019attention, li2020classifying}. Studies on facial expressions have employed machine learning to parse through facial micro-movements in images, extracting nuances of social smiles from unstructured home videos \cite{alvari2021smiling}. To explore facial dynamics, researchers estimate the steering angle of a child's head by calculating rotation parameters of 49 markers on a standardized 3D face model to classify ASD \cite{campbell2019computer}, although this method can be prone to inaccuracies due to environmental influences or camera angles. Advanced models like YOLO v3 and ResNet-18 have been used to differentiate between toddlers with ASD by examining the spatial relationship between their hands and toys \cite{qin2021vision}. Enhancements in capturing the temporal dynamics of head movements have been made through a synthesis of time-distributed Convolutional Neural Network (CNN) and Long Short-term Memory (LSTM) \cite{washington2021activity}, aimed at detecting head-banging behaviors to assist in the classification of ASD. In addition, integration of facial expression recognition, gaze estimation, and skeleton tracking has been explored to confirm the importance of grasping gestures in the supplementary diagnosis of ASD \cite{wang2019screening}. Despite these advances, a common limitation of these methods is their insufficient consideration of social interaction and cognitive behaviors, which are critical in reflecting the nuances of autism and are of considerable weight in screening and evaluation processes.

Building upon existing methodologies, different protocols have been developed to provide fresh perspectives and enhanced capabilities for the comprehensive understanding and screening of atypical behaviors in autism. The Response to Name protocol, for instance, has prompted several research groups to capture frontal facial imagery of toddlers, analyzing facial coordinates, expressions, and eye-gaze directions for screening \cite{liu2020early, washington2021activity, wang2019screening}. These groups aim to gauge engagement levels between subjects and therapists through action recognition and behavioral metrics. Yet, this protocol falls short in its ability to analyze the physical behavioral characteristics of toddlers with ASD. When it comes to fine motor skills, toddlers with ASD often display a range of hand movements that differ from those of their typically developing (TD) peers. A study designed a protocol focusing on four specific bottle grasping motions—placement, water pouring, bottle passing, and repeated water pouring—to observe these differences \cite{zunino2018video}. Other protocols, like the Expressing Needs with Index Finger Pointing (ENIFP) \cite{wang2019ENIFP}, concentrate on children's nonverbal communication abilities, specifically the actions of pointing with the index finger or palm. Nonetheless, the inherent complexity and diversity in the biological traits of ASD mean that single postural trait analyses are not sufficiently informative for an ASD diagnosis \cite{hong2020toward}, underscoring the need for more holistic approaches.

With the development of machine learning and deep learning approaches, the last decade has seen a significant emphasis on leveraging these technologies to enhance diagnostic performance. This period has witnessed the integration of wearable sensors in some studies to monitor abnormal behaviors by assessing body movement \cite{min2009detection, westeyn2005recognizing}. These sensors provide the advantages of portability and real-time physiological monitoring, although their use in clinical settings may be uncomfortable for ASD toddlers. Other research efforts have employed histograms of optical flow descriptors alongside Multi-layer Perception to pinpoint stereotypical behaviors \cite{negin2021vision}. Despite the reliance of local descriptor methods on data quality, potentially leading to cumulative errors, recent innovations have shifted towards using 3D coordinates, trajectories, and human joint positions for more accurate identification of anomalous behaviors \cite{al2020generating}. However, these newer methods often grapple with challenges in generalization and susceptibility to overfitting. In the realm of spatial feature extraction, one notable study has utilized an interactive 3D two-stream network \cite{ali2022video}, merging both RGB and optical flow modalities for classification. Another research initiative has sought to improve generalization capabilities through a guided, weakly supervised technique \cite{pandeyGuidedWeakSupervision2020a}. Additionally, the temporal dimension of behavior analysis has seen advancements with the introduction of the One Glimpse early ASD detection network \cite{tianVideoBasedEarlyASD2019a}, which combines 3D CNN with a temporal pyramid network to identify repetitive behaviors. More recently, researchers have brought together CNN and LSTM to propose a CNN–LSTM model for ASD detection with time series-based brain information \cite{xu2024autism}, reaching an classification accuracies of 81.08\% on resting state data and 74.55\% on task state data. FC-learned Residual Graph Transformer Network (RGTNet) is also proposed for ASD prediction, with an accuracy of 73.4\% \cite{wang2024residual}. While these recent advancements in deep neural networks are proficient in learning representations, they still face two major challenges: (1) the less-than-desirable prediction accuracy (often around or below 80\%), and (2) the lack of mechanisms to analyze temporal dynamics of complex actions. As such, there is a pressing need for a hybrid engineering-medical approach for ASD detection, including implementing both powerful deep learning frameworks to enhance LSTM and develop better protocols and corresponding CV information disentangling mechanisms.

\section{Methods}
In this study, we employ a video-based behavioral recognition method to evaluate the interactions between toddlers and their environment, with a specific focus on their head, hand, and block movements in a structured setting. Central to our approach is the implementation of the Parent-Child Dyads Block-Play (PCB) protocol, which is designed to assess the social attention and cognitive abilities of children. The PCB4ASD-ED dataset, a crucial component of our research, encompasses video recordings of both ASD-diagnosed and typically developing children engaged in block play with their parents. These videos have been meticulously labeled under the expert guidance of pediatric specialists from the Department of Pediatrics at Qilu Hospital of Shandong University. Complementing the PCB protocol, we have developed a novel 2sGCN-AxLSTM network. This network is adept at analyzing dynamic shifts in atypical behaviors, enabling us to establish a baseline for comparative analysis. This integrated methodological approach aims to provide a more nuanced understanding of ASD-related behaviors and enhance the effectiveness of early screening processes.

\subsection{PCB Protocol}\label{ssec:pcb}
Recent evidence underscores the substantial impact of parent-driven effects and parenting behaviors on the development of ASD. These influences differ markedly from those observed in families with TD children \cite{crowell2019parenting}. The PCB protocol, a structured sequence of tasks, is specifically designed to assess toddlers, especially those diagnosed with ASD. It evaluates parenting behaviors within a controlled experimental setting. This protocol is meticulously crafted to provide a standardized environment, facilitating the consistent assessment of social attention and cognitive abilities through interactive play. In this section, we detail the scenarios, settings, and the interactive architecture employed by the PCB protocol.

\subsubsection{Scenarios and Settings}
\begin{figure}[ht]
    \centering
	\includegraphics[width=0.6\textwidth]{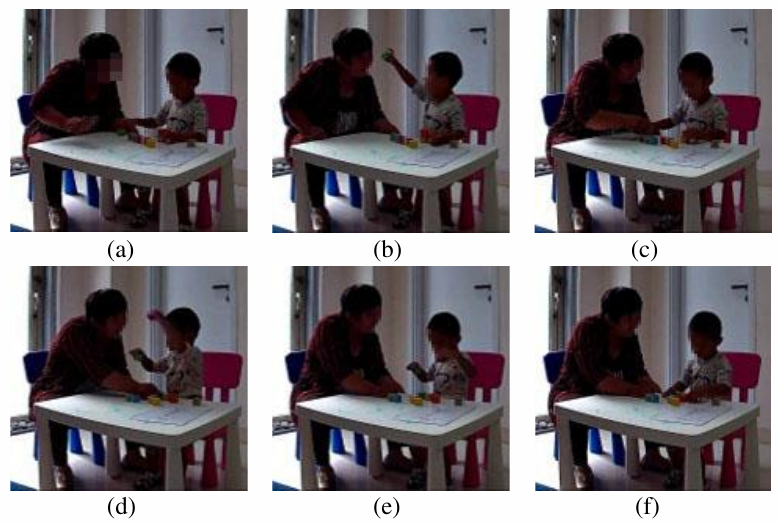}	
	\caption{Example frames of experimental scenes in the collected dataset. Faces are masked for privacy protection.} 
	\label{img:1}
\end{figure}

To evaluate the quality of the parent-child dyad activities, we design an observation platform, which is depicted in Figure \ref{img:1}. The observation room for the parent-child dyads experiment is approximately 10-15 $m^{2}$ in area and is equipped with a suitable table and two chairs. The experimental materials consist of 10 cube bricks and a box of irregular bricks for children. Before the dyads experiment begins, we provide children with the props and corresponding instructions of the task manual to help parents and children engage in the dyads.

In the observation room, a standard RGB camera with recording capability is installed to capture the facial expressions and body movements of both the toddlers and the parents. By collecting this data, we can analyze the parent-child dyads' behaviors and evaluate the quality of the interactive activities. The captured data is not limited to the current work and can be extended to other multimodal tasks, which can be explored in future research.

This observation platform provides a controlled and standardized environment for the parent-child dyads experiment. The use of standardized scene settings and instructions helps to reduce variability in the experiment and enhances the comparability of results across participants. Moreover, the use of an RGB camera with recording capability enables us to capture multiple types of behavioral data, including facial expressions, body movements, and nonverbal communication. These data can be analyzed quantitatively and qualitatively to identify patterns of parent-child dyads and evaluate the effectiveness of different intervention strategies. The use of standardized materials and instructions, as well as the collection of multimodal behavioral data, can help enhance the reliability and validity of the experiment and provide valuable insights into the development of children's social and emotional skills. 

\subsubsection{Interactive Architecture}
\begin{figure}[ht]
  \centering
  \begin{subfigure}[t]{0.38\textwidth}
    \centering
    \includegraphics[width=.4\linewidth]{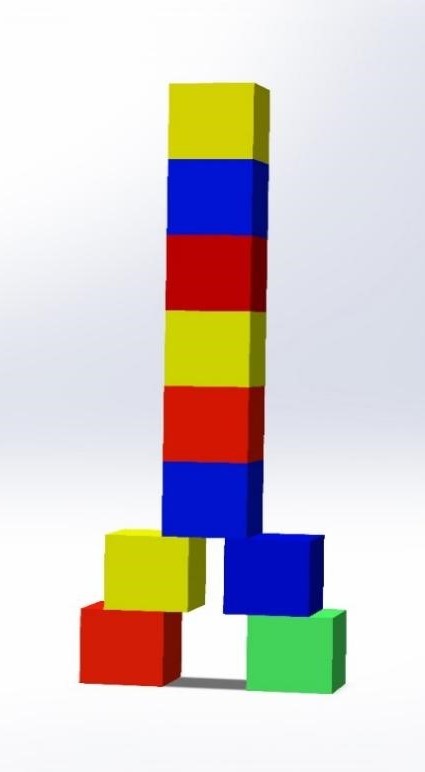}
    \caption{Example of block-play task objectives}
    \label{fig:blocks}
  \end{subfigure}
  \hfill
  \begin{subfigure}[t]{0.5\textwidth}
    \centering
    \includegraphics[width=.7\linewidth]{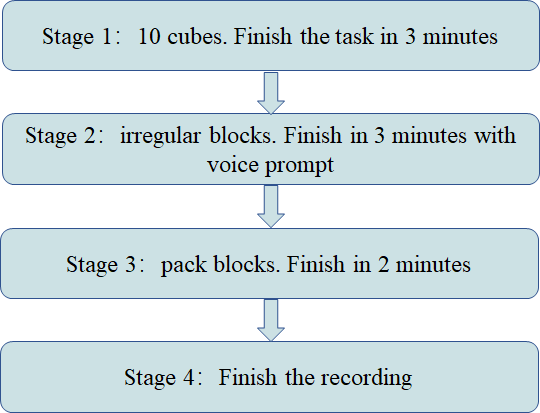}
    \caption{The workflow of PCB protocol}
    \label{fig:workflow}
  \end{subfigure}
  \caption{Parent-child interaction with props and observation of operational procedures}
\end{figure}

The process of observing parent-child dyads through completing tasks involves setting specific tasks that require collaboration between the child and the parent. The purpose of the tasks is to assess reciprocal social interaction during task completion. The observation process is conducted in three sessions, each lasting about 8 minutes. The building model of the blocks is shown in Figure \ref{fig:blocks}.

As depicted in Figure \ref{fig:workflow}, the PCB protocol is divided into four distinct stages:
\begin{itemize}
    \item Stage 1 involves providing the child and parent with ten cubes and an instruction manual to complete a set task within approximately three minutes. 
    \item In Stage 2, the observer retrieves the cubes and the instruction manual from the participants and provides the child with a box of irregular bricks to play with freely for approximately three minutes. The session begins with a voice prompt from the observer.
    \item In Stage 3, the participants are required to pack and organize the bricks used in the previous sessions for about two minutes. The session starts after an audio cue from the observer.
    \item The final stage, Stage 4, involves ending the recording process. The use of age-appropriate tasks and props promotes the participation of both the child and parent in the interactive observation. 
\end{itemize}

These stages are designed to progressively evaluate different aspects of a toddler's abilities, from simple play to following complex instructions, thereby providing a comprehensive overview of the child's behavioral patterns. 

\subsection{A Comprehensive PCB4ASD-ED Dataset}
We collect the Parent-Child Block Play for Autism Spectrum Disorder Early Detection (PCB4ASD-ED) dataset, a comprehensive dataset with more participants and duration of individual sessions than previous benchmarks. Our dataset includes 187 videos, including 97 ASD video clips and 90 TD video clips, each approximately 20 seconds long. This dataset is nearly twice as many as the previous benchmark dataset SSBD \cite{rajagopalan2014detecting} (68 videos), and there are no shared videos between these two datasets. 
\subsubsection{Data Collection and Annotation}
The PCB4ASD-ED dataset is collected by physicians in a normal examination room environment. These are continuous prolonged video behavioral data and require manually labeling every clips in a video. The dataset consists of videos of three different directive block play activities, introduced in \ref{ssec:pcb} , performed by children aged 12 months to 6 years and their caregivers. The videos are rigorously reviewed and approved for authenticity by professional clinical doctors. No real information about the subjects' conditions is included in the videos, so there are no health or pathological annotations in the dataset. The diagnosis of children with ASD is confirmed by ADOS-2, ADI-R and CARS \cite{schopler2010childhood}. The research protocol was approved by the local ethics committee (ASL3 Genovese) and was under the principles of the Helsinki Declaration \cite{world2001world}, ethical approval number KYLL-202309-044. Clinical doctors only annotated abnormal behaviors, which could potentially indicate ASD. 

Regarding the data collection devices, the parameters of the camera used to capture the videos are fixed, and the entire dataset is converted to a rate of 17 frames per second. With this conversion, there are a total of 72,418 frames in the dataset. The dataset is focused on the theme of parent-child dyads during block play and is collected from different individuals. However, there are also videos from different environments related to the same theme. There is a total of 129 participants, including 40 children with ASD and 89 typically developing children. More detailed information about the collected dataset is shown in Table \ref{tab:my_label}.

\begin{table}[ht]
	\centering
     \caption{A total of 129 annotated videos depicting children's body movements and actions are used in these experiments. A significant proportion of these sequences captured the actions performed by children during block-play activities in parent-child dyads.}
	\begin{tabular}{ccc}
		\toprule
		Details & All videos & Working subset \\
		\midrule
		NO. of videos &  185 & 187 \\
		NO. of frames &  708,744 & 72,148\\
		NO. of subjects &  129 & 129\\
		Total length of videos &  38,251.89s & 4,265.56s\\
		Average video length &  335.5s & 23.2s\\
		\bottomrule
	\end{tabular}
	
	\label{tab:my_label}
\end{table}

Among the final sample of 89 TDs  and 40 individuals with ASD, there is no significant association between gender and age in the context of behavioral screening for ASD (see Appendix A).

\subsubsection{Video Processing}
The  video frames are  processed by applying Gaussian smoothing to reduce detail in the visual appearance of the subjects. This blurs the image by giving more weight to the central pixels and less weight to the surrounding pixels, reducing noise and improving object recognition.

\textbf{Person detection.} The videos contain interactions between people and objects. Our aim here is to detect the subjects performing actions. For this, we fine-tuned a child detection algorithm using Faster R-CNN \cite{ren2015faster}, an object detection model with 13 convolution layers, 13 ReLU layers, and 4 pooling layers, to detect people in each frame. The input image is fed through a backbone network to extract feature maps, which are then used by an RPN network to generate candidate regions. Finally, the candidate regions are passed through a detection network for classification and bounding box regression. The detection network uses a RoI pooling layer to transform the candidate regions of different sizes into fixed-size feature vectors. The output includes the coordinates of the bounding boxes, object, and class predictions.

\textbf{Skeleton key points extraction.} The output of Faster R-CNN is fed into HRNet \cite{sun2019deep} (High-Resolution Network), which predicts the position of each skeleton key point for the identified human targets. HRNet has multiple parallel branches, each extracting feature at different scales using different convolution kernel sizes and strides, generating multiple scale feature maps. The different scale feature maps are then fused at the pixel and channel levels, resulting in a richer and more accurate feature representation. Finally, a fully connected layer with multiple output nodes takes the coordinates and confidence of each corresponding key point. Both models are pretrained on the COCO dataset with 80 classes. In our experiments, the network takes a $1280\times720$ tensor as input and outputs a $17\times3$ tensor.

These processing steps allow our hybrid deep learning framework to focus more on specific movements or regions of interest during behavior recognition, thereby reducing the noise level in the data and providing a more precise representation of movement patterns related to ASD and TD children.

\section{A hybrid deep learning framework}
\begin{figure*}[ht]
\centering
	\includegraphics[scale=0.4]{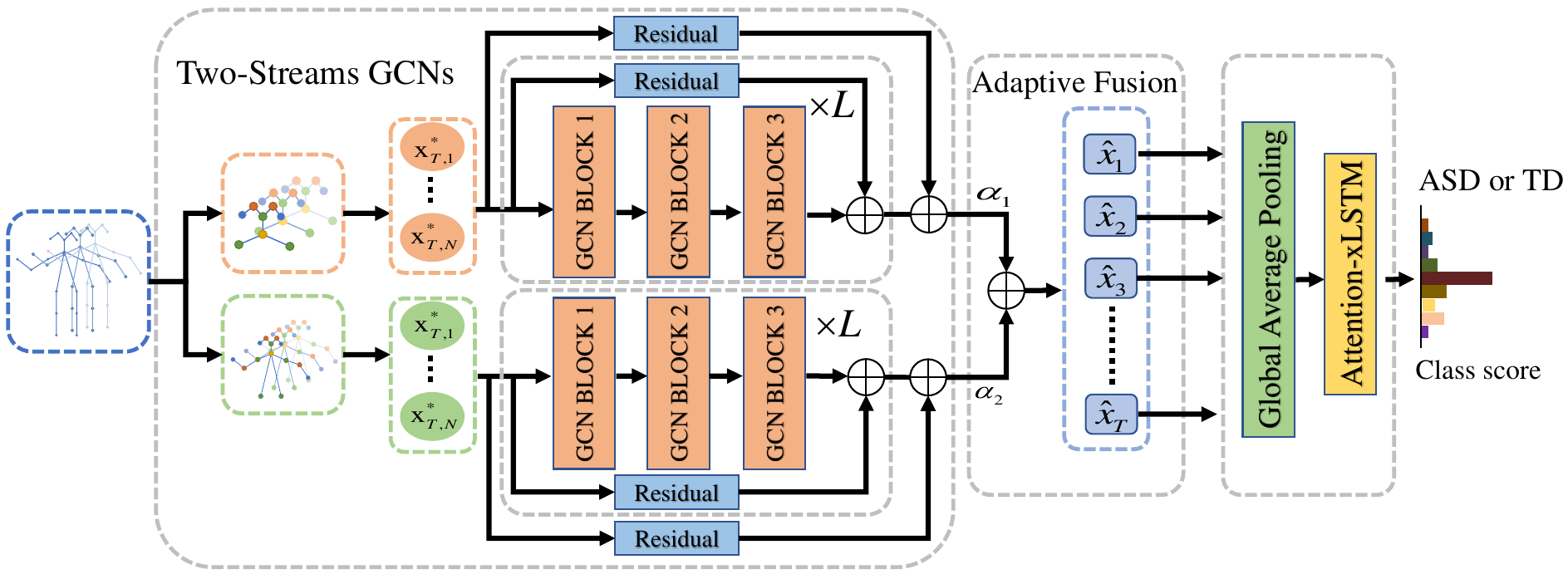}
\caption{Pipeline of the proposed 2sGCN-AxLSTM. Different graph partitions are fed to the 2S-GCNs block to learn spatial features. Then they are fused by the Adaptive Fusion module and finally fed to the xLSTM with an attention mechanism for behavioral prediction. L denotes the number of layers combined by GCN blocks.}
\label{fig:sys}
\end{figure*}

We propose a 2sGCN-AxLSTM framework, a skeleton-based action recognition method based on Graph Convolutional Networks (GCN) and Extended Long Short-Term Memory (xLSTM). GCN learns node features and network structure through the weighted aggregation of adjacent nodes, particularly effective at capturing complex spatial patterns in graph-structured data \cite{yan2018spatial}. xLSTM leverages exponential gating and an enhanced memory architecture with increased capacity, making it particularly effective for long-term time series forecasting \cite{beck2024xlstm}. Our 2sGCN-AxLSTM framework can be used as an alternative to the original GCN-based methods, surpassing GCN in accuracy under appropriate settings, while improving robustness and feature extraction in the spatiotemporal domain. Figure \ref{fig:sys} provides an overview of the 2sGCN-AxLSTM network.

\subsection{Two-Stream GCNs (2sG)}
GCN is an effective method for processing two parts of skeleton data in our task. The normalized skeleton sequence $x$ can be represented in the form of a graph $G = \{ N,E\} $ , where N is a set of joint nodes, $N = [{n_1},{n_2}, \cdots ,{n_k}]$, k is the number of nodes across T frames, and E denotes the edges connecting the nodes. Intraframe edges are defined based on the natural connections between joints, while interframe edges connect the same joints between consecutive frames.

GCNs update the features of the root node by aggregating information from local sets of spatial nodes, similar to how CNNs operate. However, because spatial graphs have a non-Euclidean structure, implementing GCNs is more complex than implementing traditional CNNs. To address this, we first define the layer-wise propagation rules for two-stream GCN (2s-GCN), using partitioning strategies and residual blocks.

\subsubsection{Data Preprocessing}

The preprocessing step of the skeleton sequence $x = \{ {x_{raw}}|t = 1, \cdots ,T\} $  with T frames involves normalization to improve stability and accelerate convergence during the training process \cite{sun2019deep} . Specifically, the original feature vector ${x_{raw}}$  of the $t - th$ frame consists of two sub-vectors $\{ {{\rm{x}}_{raw,m}}|m = 1,2\} $ , representing different body parts (e.g. upper body and head). These sub-vectors are labeled based on their part membership, where the label variable m is set to 1 for the upper body and 2 for the head. To ensure that each sub-vector is independently normalized, the normalization process is performed on ${x_{raw,m}}$ , which is expressed mathematically as:
\begin{flalign}
     {x_{raw,m}} = \frac{{{x_{raw,m}} - {{\bar x}_{raw,m}}}}{{\sigma ({x_{raw,m}})}}
\end{flalign}
where  ${\bar x_{raw,m}}$ and $\sigma ({x_{raw,m}})$  are the mean and standard of ${x_{raw,m}}$  respectively.

\subsubsection{Partitioning Strategy}

The partitioning strategy under consideration is restricted to skeleton data characterized by a complex topology in the spatial dimension. Specifically, ${G_t} = \{ {N_t},{E_t}\} $  represents the spatial graph of the skeleton in frame t, where the neighbor set of the root node ${v_{ti}}$  is designated as $N({v_{ti}}) = \{ {v_{tj}}|d({v_{ti}},{v_{tj}}) \le 1\} $ . Here, i and j represent the node labels, while $d({v_{ti}},{v_{tj}})$  corresponds to the minimum path length from ${v_{ti}}$  to  ${v_{tj}}$. Notably, distinct neighbor sets may exhibit variations in the number and order of nodes, rendering direct implementation of kernel sharing unfeasible. To overcome this challenge, two partitioning strategies have been devised to partition the neighbor set into a fixed number of K subsets, accomplished by assigning the label  $\{ 1, \cdots ,K\} $ to each node  ${v_{tj}} \in N({v_{ti}})$ via the mapping function ${l_{ti}}:N({v_{ti}}) \to \{ 1, \cdots ,K\} $ .

One strategy for partitioning a set of neighbors is based on the distance between each node and a designated root node, which is mainly used for critical points in the head. This method, known as distance partitioning, involves dividing the set of neighbors into subgroups based on the minimum path length from each interior node to the root node. Formally, the distance partitioning can be expressed as the equation:
\begin{flalign}
    {l_{ti}}({v_{tj}}) = d({v_{ti}},{v_{tj}}) + 1
\end{flalign}
Where ${l_{ti}}({v_{tj}})$ represents the label of node ${v_{tj}}$  in $N({{\rm{v}}_{ti}})$ . This partitioning method involves dividing the set of neighbors of a node into two distinct subsets, namely the root node and its 1-neighbor nodes. 

\begin{figure*}[ht]
\centering
	\includegraphics[scale=0.8]{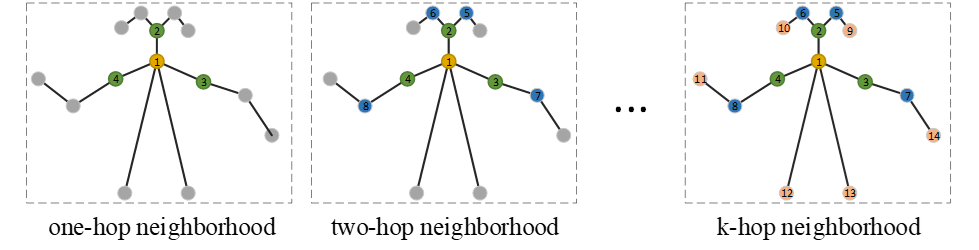}
\caption{Example of different neighborhoods of a node.}
\label{fig:skele}
\end{figure*}

The other strategy, namely multi-scale spatial partitioning, addresses the issue of biased weights of neighboring nodes that are relatively far apart, as shown in Figure \ref{fig:skele}. Node self-loops in GCNs can lead to more possible loops, which may amplify biases and result in skeleton key point sequences being dominated by signals from local body parts. Self-cycling can also cause the model to fail in capturing long-range key point dependencies of high-order polynomials. To overcome this issue, different k-values are assigned to different adjacency matrices to obtain different scales. We apply this allocation method to the skeleton key points of the upper body. This is achieved mathematically by writing the equation as:

\begin{flalign}
    {l_{ti}}({v_{tj}}) = \left\{ {\begin{array}{*{20}{c}}
1\\
1\\
0
\end{array}\begin{array}{*{20}{c}}
{}&{ifd({v_i},{v_j}) = k,}\\
{}&{ifi = j}\\
{}&{otherwise,}
\end{array}} \right.
\end{flalign}

\subsubsection{Graph Convolution Implementation}
The spatial aggregation strategy for graph convolution can usually be expressed mathematically as:
\begin{flalign}
    {Y_{out}}({v_{ti}}) = \sum\limits_{{v_{tj}} \in N({v_{ti}})} {\frac{1}{{{Z_{ti}}({v_{tj}})}}X({v_{tj}})} W({l_{ti}}({v_{tj}}))
\end{flalign}
where $X({v_{tj}})$  is the input feature of node ${v_{tj}}$ .  $W( \cdot )$ is a weight function, allocated by the label  ${l_{ti}}({v_{tj}})$ from K weights. ${Z_{ti}}({v_{tj}})$ is the number of neighbor ${v_{tj}}$ , which normalizes feature representations. ${Y_{out}}({v_{ti}})$  denotes the outputs of the graph convolution layer at node ${v_{tj}}$ . To be more specific, the adjacency matrix A of the skeleton graph with self-loops can be divided into K matrices $\{ {A_k}|k = 1, \cdots ,K\} $  based on partition strategy. Mathematically, this can be expressed as $A = \sum\nolimits_k {{A_k}} $ . To illustrate, both distance partitioning and spatial configuration partitioning can be expressed as ${A_1} = I$  where I is the identity matrix. Similarly, the degree matrix  $\Lambda $ can also be dismantled into K matrices $\{ {D_k}|k = 1, \cdots ,K\} $ , following the same partitioning strategy. The Eq.\ref{eq:5} can be represented as:
\begin{flalign}
    {Y_{out}} = \sigma (\sum\limits_{k = 1}^K {\Lambda _k^{ - \frac{1}{2}}{A_k}\Lambda _k^{ - \frac{1}{2}}X{W_k}} )
    \label{eq:5}
\end{flalign}

where  $\sigma $ denotes an activation function.  $\Lambda _k^{ - \frac{1}{2}}{A_k}\Lambda _k^{ - \frac{1}{2}}$ is the normalized k-adjacency by Symmetric Normalization \cite{kipf2016semi}.

In the strategy of multi-scale spatial partitioning, we define a new adjacency matrix $\hat A$ , the following equation is obtained in Eq.\ref{eq:5}.
\begin{flalign}
    \hat A = \Lambda _k^{ - \frac{1}{2}}({A_k} + I)\Lambda _k^{ - \frac{1}{2}}
\end{flalign}
\begin{flalign}
    \hat A_k^{} = \min ({(\Lambda _k^{ - \frac{1}{2}}({A_k} + I)\Lambda _k^{ - \frac{1}{2}})^k},1)
\end{flalign}
\begin{flalign}
    {Y_{out}} = \sigma (\sum\limits_{k = 1}^K {{{\hat A}_k}X{W_k}} )
\end{flalign}

where min is the minimum function. Based on (5) and (6), the formula of 2S-GCN on the whole input feature map $X = {R^{N \times T \times C}}$ , where N, T, and C are the number of joints, frames, and channels.

\begin{figure}[ht]
\centering
	\includegraphics[scale=0.5]{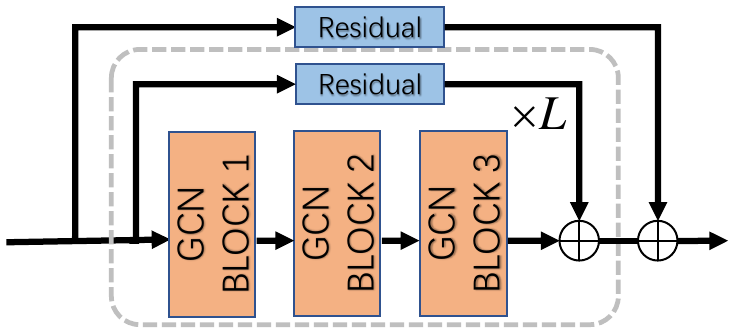}
\caption{The GCN Block Architecture. The complete network acquires knowledge about the residual vectors between the input and target pose sequences through the incorporation of global residual connections.}
\label{fig:blk}
\end{figure}

To extract features from pose data, the GCN module first maps the input from pose space to feature space. Then, GCN blocks are utilized to extract features in this feature space, with residual connections added between every three GCN blocks. The L used in this experiment is 3, i.e. a total of 9 GCN blocks are used. The network module is then able to learn the residuals instead of the target pose directly. Finally, residual connections are added between the input and output poses, ensuring that the network learns the differences between them. This residual connection is intended to improve the accuracy of pose feature extraction. The GCN Block Architecture is shown in Figure \ref{fig:blk}.

\subsubsection{Adaptive Fusion}
We utilize an adaptive fusion module to assign weights for the fusion of multiple features in our ASD parent-child dyads block-play task. Although it’s common in many studies to add or splice multimodal features, the arm's role is notably more significant than the head's secondary role in our task. As a result, we designed a weight assignment mechanism that considers this hierarchical relationship between features.
Let ${\hat x_{t,m}}$  denotes the 256-dim feature vector obtained by mapping the output of the m-th part in the t-th frame from a Multilayer Perceptron. The formula is used to fuse the multi-feature of the t-th frame and can be expressed mathematically as: 
\begin{flalign}
    {\hat x_t} = \sum\limits_m {{\alpha _m}{{\hat x}_{t,m}}} 
\end{flalign}
where ${\alpha _m}$  refers to the spatial importance weights assigned to the part of label m, which are adaptively learned by the network. Inspired by \cite{wang2019adaptively}, we restrict $\sum\limits_m {{\alpha _m}} $ to 1 and ${\alpha _m} \in [0,1]$, and define 
\begin{flalign}
    {\alpha _m} = \frac{{\exp (\lambda {\omega _m})}}{{\sum\nolimits_{n = 1}^M {\exp (\lambda {\omega _n})} }}
\end{flalign}
where $\lambda $ is a reinforcement factor that is used to control variation amplitude in $\alpha $. $\omega $ is a set of parameters used to iteratively optimize the model, initialized with zero. And $\omega $ can be learned through standard back-propagation.

\subsection{Attention Enhanced xLSTM (AxLSTM)}

The information provided by frames within a skeleton sequence is not equally valuable. Keyframes contain the most distinguishing information, while other frames provide contextual information. In the ASD parent-child dyad during a ``block stacking`` action, the ``hand approaching`` sub-phase is considered more important than the ``arm opening`` sub-phase. To address this issue of unequal importance in frames, we develop an attention enhances xLSTM module, namely AxLSTM, based on the vision-specific implementation Vision-LSTM (ViL). 

\subsubsection{xLSTM and Vision-LSTM (ViL)}




LSTM \cite{hochreiter1997long}, a variant of RNN, exhibits remarkable capability in modeling long-term dependencies for sequence modeling. The LSTM architecture includes three gates: an input gate \(i_t\), a forget gate \(f_t\), and an output gate \(o_t\). These gates control the flow of information through the cell state \(C_t\) and the hidden state \(h_t\) as follow:

\begin{flalign}
\begin{array}{l}
    i_t = \sigma(W_i x_t + U_i h_{t-1} + b_i)\\
    f_t = \sigma(W_f x_t + U_f h_{t-1} + b_f)\\
    o_t = \sigma(W_o x_t + U_o h_{t-1} + b_o)\\
    C_t = f_t \odot C_{t-1} + i_t \odot \tanh(W_c x_t + U_c h_{t-1} + b_c)\\
    h_t = o_t \odot \tanh(C_t)
\end{array}
\end{flalign}

\begin{figure}[ht]
    \centering
    \begin{subfigure}[b]{0.45\textwidth}
        \centering
        \includegraphics[width=\textwidth]{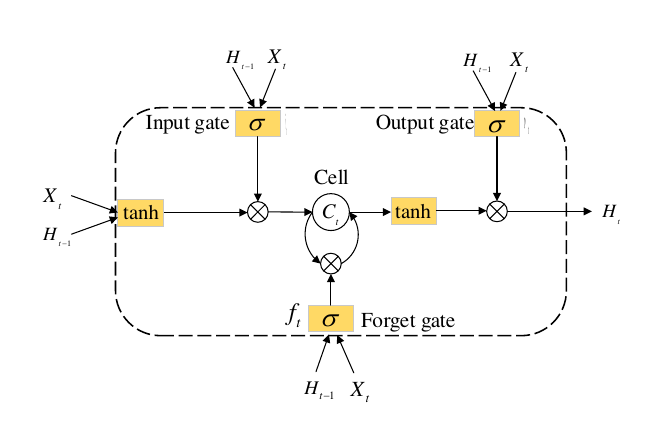}
        \caption{The LSTM architecture}
    \end{subfigure}
    \hfill
    \begin{subfigure}[b]{0.50\textwidth}
        \centering
        \includegraphics[width=\textwidth]{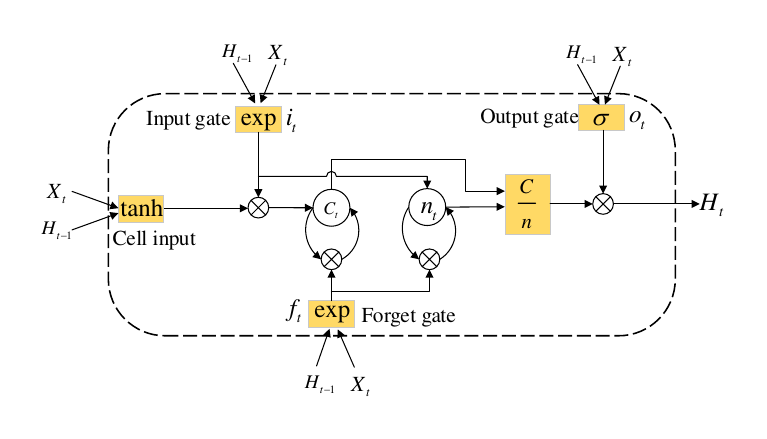}
        \caption{The sLSTM with an enhanced gating mechanism}
    \end{subfigure}
    \caption{Comparison of sLSTM and traditional LSTM structures}
    \label{fig:lstm}
\end{figure}

Here, \(x_t\) is the input at time \(t\), \(W\) and \(U\) are weight matrices, \(b\) is the bias term, and \(\sigma\) is the sigmoid activation function. The structure of the LSTM is shown in Figure \ref{fig:lstm}.

The xLSTM extends the traditional LSTM architecture to address challenges in handling large-scale data and long sequences, such as poor parallelism and limited storage capacity. It introduces an enhanced gating mechanism and a novel matrix memory structure to overcome these limitations \cite{beck2024xlstm}. Specifically, the xLSTM uses exponential gating and a parallelizable matrix memory, allowing the model to efficiently handle longer sequences with improved stability and scalability:

\begin{flalign}
\begin{array}{l}
\text{Gate}_t = \sigma(W_g x_t + U_g h_{t-1} + b_g)\\
\text{Memory}_t = \text{Gate}_t \odot \text{Memory}_{t-1} + (1 - \text{Gate}_t) \odot \tanh(W_m x_t + U_m h_{t-1} + b_m)\\
h_t = \text{Memory}_t \odot \tanh(W_h x_t + U_h h_{t-1} + b_h)
\end{array}
\end{flalign}
These improvements significantly boost the performance and scalability of LSTM networks for large-scale data applications, as demonstrated in \cite{beck2024xlstm}.

The sLSTM unit of xLSTM, which is the key component in Vision-LSTM (ViL) \cite{alkin2024vision}, enhances traditional LSTM by adding a scalar update mechanism, offering finer control over memory cells and improving gating for sequence data with subtle temporal variations. The sLSTM uses exponential gating and normalization to achieve stability and accuracy in long-sequence processing:
\begin{flalign}
\begin{array}{l}
\text{ScalarUpdate}_t = \text{exp}\left(\frac{W_s x_t + U_s h_{t-1} + b_s}{\text{Norm}(W_n x_t + U_n h_{t-1} + b_n)}\right)\\
h_t = \text{ScalarUpdate}_t \odot \tanh(C_t)\\
\end{array}
\end{flalign}
These enhancements allow sLSTM, and ultimately ViL, to match the performance of more complex models with lower computational costs, excelling in classification tasks such as those on the PCB4ASD-ED dataset \cite{alkin2024vision}.

\subsubsection{AxLSTM}

In AxLSTM, the temporal attention mechanism learns different attentional weights for different frames $\beta$ and captures more accurately in the sequence. It is able to efficiently correct the decision making when a new context is encountered, and, at the same time, increases the ability of parallel computing. The cell memory ${c_t}$ exhibits temporal dynamics through its weight as a self-connected recursive edge, and in conjunction with the hidden state ${H_t}$ . The functions of the sLSTM unit are defined as follows:
\begin{flalign}
    \begin{array}{l}
{i_t} = \exp (W_{xi}^{}{X_t} + {W_{hi}}{H_{t - 1}} + {b_i})\\
{f_t} = \sigma ({W_{xf}}{X_t} + {W_{hf}}{H_{t - 1}} + {b_f}) \;\text{OR}\; \exp({W_{xf}}{X_t} + {W_{hf}}{H_{t - 1}} + {b_f})\\
{o_t} = \sigma ({W_{xo}}{X_t} + {W_{ho}}{H_{t - 1}} + {b_o})\\
{u_t} = \tanh ({W_{xc}}{X_t} + {W_{hc}}{H_{t - 1}} + {b_c})\\
{C_t} = {f_t} \odot {C_{t - 1}} + {i_t} \odot {u_t}\\
{N_t} = {f_t} \odot {n_{t - 1}} + {i_t}\\
{H_t} = {o_t} \odot \tanh\left(\frac{C_t}{N_t} \right)
\end{array}
\end{flalign}

where $ \odot $  denotes the Hadamard product.  $\sigma $ is the sigmoid activation function. ${u_t}$ is the modulated input.

\begin{figure}[ht]
\centering
	\includegraphics[scale=0.5]{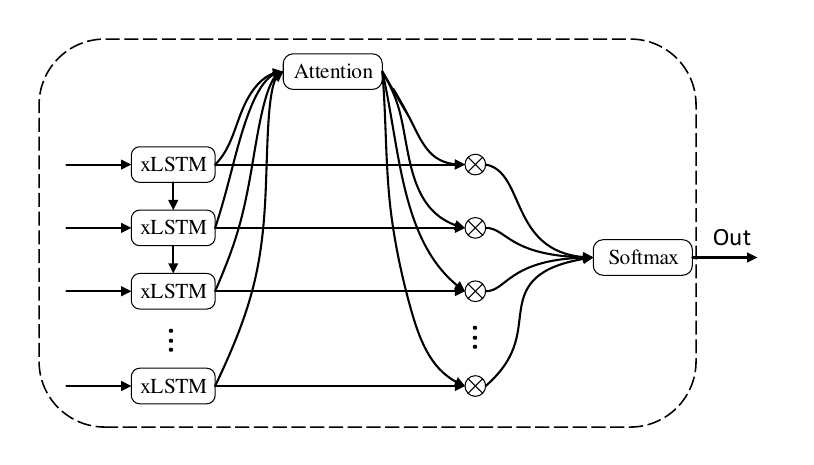}
\caption{Structure of the attention-enhanced xLSTM mechanism}
\label{fig:alstm}
\end{figure}

The attention layer takes the hidden state \quad $h = {[{h_k},{h_{k + 1}}, \cdots , {h_{k + w - 1}}]^T}$, \quad${h_i} \in {R^{1 \times n}}$  as its input, and based on this input, computes a set of attention weights ${\alpha _k},{\alpha _{k + 1}}, \cdots ,{\alpha _{k + w - 1}}$ that represent the magnitude of the influence of each hidden state on the outcome. The inputs are then weighted and summed by the model to obtain the resulting vector ${l_k}$. The attention layer's architecture is illustrated in Figure \ref{fig:alstm}. To enhance the model's performance, this attention mechanism allows the model to focus more on the important parts of the input sequence and less on the less relevant ones.

The attention mechanism can be represented by the following equation:
\begin{flalign}
    {W_a} = W\left[ {\begin{array}{*{20}{c}}
{{x_t}}&{{h_{t - 1}}}
\end{array}} \right] + b
\end{flalign}
\begin{flalign}
  {e_i} = \frac{1}{n}\sum\limits_j {{w_{i,j}}} 
\end{flalign}

\begin{flalign}
    {\alpha _j} = \frac{{\exp ({e_j})}}{{\sum\limits_{i = k}^{k + w - 1} {\exp ({e_j})} }}
\end{flalign}
\begin{flalign}
    {l_k} = {\mathop{\rm Re}\nolimits} LU(\sum {{\alpha _i}{h_i}} )
\end{flalign}

where ${W_a} \in {R^{w \times n}}$  denotes the weight matrix. ${w_{i,j}}$  denotes the element in the matrix ${W_a}$ . $b \in {R^{w \times n}}$ and  $W \in {R^{w \times n}}$ are learnable parameters. ${l_k}$  is the result vector. In the process of model training, the model inherently learns the impact of each input element on the output and generates attention weights for each time step. As the sliding window shifts, the input sequence values change, but the attention layer can calculate attention weights based on input values, enabling the model to flexibly focus on changes in the input values. This approach facilitates the model in capturing the critical information in the input sequence more precisely, thereby enhancing model performance.

\section{Results}
In this section, we present the results of ablation studies, which systematically assess the impact of each component utilized in 2sGCN-AxLSTM. We also provide performance comparison between our method and contemporary ASD screening methods, namely RGB-based, pose-based, and CNN-based approaches.

\subsection{Ablation Study of Motor Development in Children with ASD}
\begin{figure}[ht]
    \centering
    \includegraphics[width=3.5in]{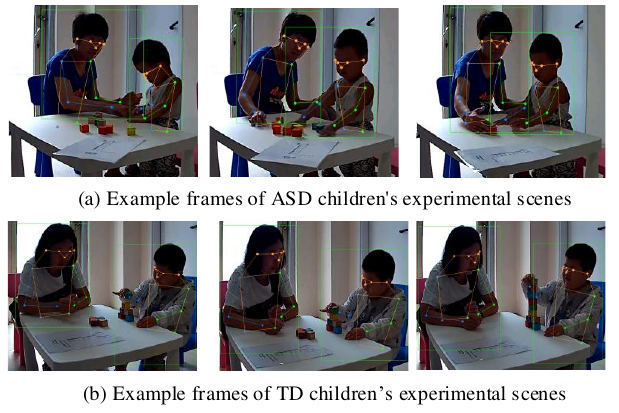} 
   
  \caption{Video clips of the experimental scenes. Children with ASD struggle to build blocks smoothly, whereas children with TD can do so with relative ease. Faces are covered by annotation for privacy protection.}
  \label{fig:scenes}
\end{figure}

By analyzing the behavior of subjects involved in block play activities, we can evaluate their ability to construct block structures under parental guidance. Figure \ref{fig:scenes} illustrates the different interactions between children and their parents during block building at different time periods. Figure \ref{fig:scenes}(a) shows that children with ASD frequently require assistance from their parents and struggle to meet the task requirements. In contrast, Figure \ref{fig:scenes}(b) demonstrates that children with TD are capable of independently complete the block building on thier own. The experimental results show high precision in the identification of people diagnosed with ASD. Our research used the PCB protocol to examine motor development in children with ASD, with a particular focus on head and upper limb interaction behaviors. These include head turning and physical contact between children, their parents, and objects such as blocks. To ensure a reliable and objective observation of subjects' interaction abilities, we use a standardized scenario and structured assessment. In this case, we focus on upper body movements and head movements and obtain results with an accuracy of 0.89 and an unweighted average recall of 0.85 (Table \ref{tab:3}). 

Table \ref{tab:3} first shows the screening results with different physical characteristics. The worst performance is achieved using a single feature of Class 1 (head) or Class 2 (upper body), likely due to the low number of features detected using a single feature. As shown in the “Fuse Features” column of Table \ref{tab:3}, there are a recognizable performance improvement of 3$\%$ (from 0.76 to 0.79) when using the adaptive fusion module, compared to the best performing result between the two classes (class 2). This model performance improvement demonstrates the effectiveness of fusing information from multiple feature maps, thereby avoiding extraneous data. Moreoever, as shown in the ``With Attention'' column of Table \ref{tab:3}, the further inclusion of the attention mechanism in the model leads to a substantial performance improvement of 13$\%$ (from 0.76 to 0.79), showcasing the outstanding feature recognition ability of our 2sGCN-AxLSTM framework. 

\begin{table}[ht]
    \centering
        \caption{Accuracy of video classification: each category is experimented with complete data. Class 1 and Class 2 are videos with head movements and upper body behaviors, respectively. The highest accuracy and UAR are achieved when ``full data'' and ``with attention'' are presented, marked in bold.}
            
   \begin{tabular}{ccccccc}
\toprule
                                             &          & Class1 & Class2 & Fuse Features & With Attention & Full data \\ \midrule
\multirow{2}{*}{ASD data}                    & Accuracy & 0.64   & 0.71   & 0.76          & 0.77           & 0.79      \\
                                             & UAR      & 0.63   & 0.69   & 0.73          & 0.75           & 0.75      \\
\multirow{2}{*}{TD data}                     & Accuracy & 0.59   & 0.62   & 0.64          & 0.67           & 0.70      \\
                                             & UAR      & 0.56   & 0.60   & 0.62          & 0.65           & 0.69      \\
\multirow{2}{*}{Full data}                   & Accuracy & 0.77   & 0.81   & 0.85          & \textbf{0.89}  & /         \\
                                             & UAR      & 0.74   & 0.79   & 0.84          & \textbf{0.85}  & /         \\ \bottomrule
\end{tabular}
            
  \label{tab:3}
\end{table}

\begin{table}[ht]
    \centering
    \caption{Evaluation results for all frameworks: local descriptor-based approach, articulated pose-based approach, CNN-based approach, and our approach.}
    \begin{tabular}{@{\extracolsep{\fill}}rcc} 
        
        \toprule
        & Accuracy & Unweighted Average Recall \\ 
        \bottomrule
        \multicolumn{1}{l}{\textit{Local Descriptor RGB-based method:}} &          &                           \\
        HOF-BOVW                   & 0.63     & 0.67                      \\
        \multicolumn{1}{l}{\textit{Pose-based methods:}}         &          &                           \\
        STGCN                      & 0.62     & 0.64                      \\
        Skeleton-LSTM              & 0.68     & 0.67                      \\
        \multicolumn{1}{l}{\textit{CNN-based methods:}}          &          &                           \\
        3DCNN                      & 0.44     & 0.46                      \\
        PoseC3D                    & 0.75     & 0.73                      \\
        \multicolumn{1}{l}{\textbf{\textit{Our methods:}}}                 &          &                           \\
        2sGCN-ALSTM             & 0.79          & 0.78                      \\
        \textbf{2sGCN-AxLSTM}                  & \textbf{0.89}     & \textbf{0.85}                      \\ 
        
        \bottomrule
        
    \end{tabular}
    
    \label{tab:4}
\end{table}

In addition, as Table \ref{tab:3} indicates, the attention mechanism enhances the model's capacity to capture long-distance frames' behavioral features and overcomes performance degradation associated with long-sequence analysis. The three rows labeled ``ASD data,'' ``TD data,'' and ``Full date`` in Table \ref{tab:3} demonstrate that models trained using only ASDs or TDs as samples exhibit poor performance, with an accuracy of 0.77 and 0.67 respectively, while, when tested with Full data (comprising both ASD and TD data), our model reaches its best performing accuracy of 0.89. These findings show the necessity of augmenting the diversity of samples and models relying on a single symptomatic sample that has suboptimal generalization performance.

To summarize, our results in the ablation study suggest that children with ASD show significant differences in block-play behavior compared to TD children. Specifically, toddlers with ASD lacked fluency and naturalness in their behavioral interactive movements, especially when building block structures. In addition, there are significant differences in their ability to complete block-play tasks with parental guidance and they exhibit less interactivity compared to their TD peers.

\subsection{Comparison with Contemporary ASD Screening Methods}
In this task, children with ASD had greater difficulty completing the task compared to TD children. This is evidenced by three key behaviors: first, a tendency to arrange the blocks in rows rather than forming the required shape; second, frequent arm-waving that disrupted the tabletop environment; and third, increased likelihood of parental assistance for children with ASD compared to TD children. To gain a more comprehensive understanding of the formidable nature of the proposed task, we conduct a series of experiments by testing various ASD methodologies documented in the previous action recognition literature. Specifically, in the Local Descriptor-based method, we employ the use of Histogram of Optical Flow (HOF) descriptors with default parameters, followed by subjecting our data to the Bag of Visual Words (BoVW) framework for encoding. Moreover, we fine-tune two skeleton feature-based methods, namely STGCN \cite{kipf2016semi} and Skeleton-LSTM \cite{nie2017monocular}. Furthermore, to extend the breadth of our proposed network comparison, we also evaluate our dataset by fine-tuning two Convolutional Neural Network (CNN)-based networks, namely 3DCNN and PoseC3D \cite{duan2022revisiting}. We also compare our method, which is based on xLSTM, to the version based on LSTM, both enahnced with attention, demonstrating the unique contribution of xLSTM in our framework. 

The evaluation results of all developed and fine-tuned frameworks are presented in Table \ref{tab:4}, which is categorized into four classes: local descriptor-based methods, pose-based methods, CNN-based methods, and our methods. For training the 3DCNN, we employ the cross-validation method in tandem with Grid Search to optimize the parameters of the model, ultimately selecting a batch size and epoch of 8 and 64, respectively. In contrast, PoseC3D yields superior results by fine-tuning the network parameters, which are found to be most effective with a batch size of 16 and 24 epochs. Notably, the skeleton-based approach involves initially processing the Faster-RCNN preprocessed image and subsequently extracting the joints from the image using OpenPose. Across all categories, our best performing approach, 2sGCN-AxLSTM, outperforms all other methods and reaches an accuracy of 0.89, proved to be more effective in preserving spatial-temporal features in our dataset. 


\section{Discussion}
Our study marks a significant advancement in the early detection and screening of Autism Spectrum Disorder (ASD) in toddlers. The findings underscore the unique behavioral patterns of children with ASD during block play, particularly in their interactive movements and task completion with parental guidance, which are less fluid and natural compared to typically developing (TD) children. 

The introduction of the Parent-Child Dyads Block-Play (PCB) protocol represents a substantial contribution, establishing a controlled environment to consistently evaluate social attention and cognitive abilities through structured play. The comprehensive PCB4ASD-ED dataset, with its greater participant count and extended session durations, provides a robust basis for fine-grained behavioral analysis, far exceeding the scope of previous datasets like the SSBD. This wealth of data has been pivotal in training our hybrid deep learning framework, the 2sGCN-AxLSTM, which has demonstrated superior performance in action recognition, robustness, and feature extraction within the spatiotemporal domain. Notably, our framework outperforms conventional GCN-based and CNN-based methods, as well as the traditional LSTM-based method we developed, highlighting its effectiveness in preserving and analyzing spatial-temporal features unique to ASD behaviors. These collective advancements offer promising avenues for improving ASD diagnostic tools and therapeutic interventions.

\subsection{Limitations}


While this study marks a significant advancement in ASD detection, it also brings to light several limitations that must be addressed to enhance the robustness and applicability of the proposed methods. The insights derived from the PCB protocol and the 2sGCN-AxLSTM framework underscore the potential for integrating these technologies into clinical practice. However, the variability of ASD symptoms poses a substantial challenge, necessitating the use of larger and more diverse datasets to refine the accuracy and generalizability of these methods.

One of the primary limitations lies in the reliance on behavioral analysis videos for autism screening. Although these videos provide valuable data, they are subject to the inherent heterogeneity of autism, which can vary widely across individuals. This variability complicates the development of universally applicable models and highlights the need for more extensive, labeled datasets to capture the full spectrum of ASD manifestations.

Moreover, the use of video data introduces specific technical challenges, particularly regarding the occlusion of body parts during parent-child interactions. For example, a parent's arm may inadvertently obscure the child's hand or face, leading to inaccuracies in pose estimation and, consequently, in the behavioral analysis. Addressing these occlusion issues is critical for improving the precision of the proposed framework.

Another significant challenge is the scarcity of large-scale, labeled medical data, which is crucial for training robust machine learning models. The current limitations in data availability restrict the ability to generalize findings across diverse populations, potentially limiting the broader applicability of the research outcomes.

Moving forward, addressing these limitations will be essential for advancing the field of ASD detection. Efforts to diversify datasets, improve pose estimation accuracy in occluded scenarios, and develop models that can account for the heterogeneity of ASD will be key areas of focus. By tackling these challenges, future research can build on the foundations laid by this study and contribute to more accurate, reliable, and inclusive ASD detection methods.

\subsection{Research Outlook}
Advancing early detection methods for Autism Spectrum Disorder (ASD) requires the integration of more modality-rich datasets to enhance data diversity and robustness. However, the time-consuming and costly nature of medical data collection and labeling makes it impractical for a single institution to gather the necessary volume of data. To overcome this, self-supervised learning methods offer a promising solution by enabling the extraction of valuable information from limited datasets with minimal supervision. This approach can effectively address the constraints of scarce data, as evidenced by recent research demonstrating its potential to maximize data utility with limited supervision \cite{chen2023self}. However, managing and analyzing this multimodal data will present new challenges, particularly in isolating specific actions within continuous video streams. Addressing these challenges will be a key priority in our future research, as we work to confirm the feasibility and reliability of video-based, computer-assisted applications in healthcare and medical diagnosis.

Furthermore, ensuring the efficiency and privacy of medical data usage is crucial, particularly given the sensitive nature of such data. Innovative privacy-preserving techniques have been shown to mitigate the risk of sensitive data leakage, making them essential for safe data analysis and sharing \cite{yang2023invertible}. These techniques will be incorporated into our future work to protect patient confidentiality while enabling robust data analysis. In tackling the challenge of multimodal data fusion, recent frameworks have proven effective in integrating data from various modalities, thereby facilitating a more comprehensive analysis of ASD \cite{kim2024self}. Implementing such frameworks will be a key focus in our future research, as they are critical for harnessing the full potential of diverse datasets.

Additionally, inspired by the complexity and variability of actions observed in ASD, we plan to enhance our current framework by incorporating a multiple visual fields platform. This approach will allow us to capture and analyze the nuanced behaviors exhibited in ASD, thereby supporting the development of personalized rehabilitation strategies. Personalized interventions, tailored to the specific needs of individuals, have the potential to significantly improve treatment outcomes \cite{ke2024two}. To ensure the practical applicability of our methods, collaboration with domain experts will be essential. Their expertise will guide the validation and refinement of our approach, ensuring that it aligns with real-world clinical practices.

\section{Conclusions}
Early diagnosis of neurobiological disorders is crucial for optimal care and effective treatment. Therefore, individuals diagnosed with ASD can greatly benefit from interventions based on the level of impairment. However, this diagnostic process requires long-term monitoring and assessment of self-stimulatory behaviors during child interactions, making early diagnosis of ASD challenging. In this study, we propose a PCB4ASD-ED dataset collected from the process of block building during parent-child dyads. A behavior recognition framework is developed to identify these characteristic behaviors using 2sGCN-AxLSTM. The framework takes in video recordings and describes each behavior through extracted skeleton joint information. The child and parent skeleton joints are input into two graph streams, and an Attention-xLSTM is designed to enhance the extraction of global contextual and temporal motion features. The results reveal that the behavior recognition framework using two streams GCN and attention-based LSTM outperforms other methods and achieves competitive performance with an accuracy of 89.6\%. 

The implications of this study are significant for advancing the early diagnosis and intervention strategies for ASD. By introducing the PCB4ASD-ED dataset and an innovative behavior recognition framework, this research addresses the challenges of detecting self-stimulatory behaviors in naturalistic parent-child interactions, which are often difficult to assess in clinical settings. The use of advanced techniques like 2sGCN-AxLSTM to analyze video recordings and extract meaningful behavioral patterns from skeletal joint information demonstrates the potential for more accurate and timely ASD diagnosis. This approach not only enhances the understanding of child behavior during interactions but also provides a robust tool that could be integrated into diagnostic processes, potentially leading to earlier and more personalized interventions for children with ASD. The framework's competitive performance suggests that it could serve as a foundation for future developments in automated behavior analysis, ultimately contributing to better clinical outcomes and improved quality of life for individuals with ASD.

\section*{ETHICS STATEMENT}
The authors are accountable for all aspects of the work in ensuring that questions related to the accuracy or integrity of any part of the work are appropriately investigated and resolved. The study was approved by the Research Ethics Board at Qilu Hospital of Shandong University and was confirmed to comply the Declaration of Helsinki, ethical approval number KYLL-202309-044. Prior to the experiment, instructions about the experiment were provided, and consent was obtained from the guardians of all participating children.

\section*{Acknowledgement}
This work was jointly supported by the Key Development Program for Basic Research of Shandong Province under Grant Number: ZR2019ZD07; The National Natural Science Foundation of China-Regional Innovation Development Joint Fund Project under Grant Number: U21A20486; The Fundamental Research Funds for the Central Universities under Grant Number: 2022JC011.


\section*{Appendix A. The demographic information}

In the 129 ASD behavioral screening sample, there was no significant association between gender and age, which did not affect the experimental results as shown in Table \ref{tab:demographics}. Following statistical analysis, the chi-square statistic for gender among the subjects' toddlers yields a value of 0.594, accompanied by a p-value of 0.462. Additionally, the independent sample t-test for age results in a statistic of -0.07, with a corresponding p-value of 0.945. Both statistics exceed the statistical significance threshold of 0.05. The statistical analyses are carried out using SPSS Version 26 (IBM, Armonk, NY, USA).

\begin{table}[H]
    \centering
    \caption{Subject demographics and group comparisons}
    \begin{tabular}{ccccc}
        \toprule
        & \makecell[c]{ASD\\N($\%$)} & \makecell[c]{TD\\N($\%$)} & \makecell[c]{Group\\comparison} &  \makecell[c]{P value} \\
        \midrule
        No. of Subjects & 40(31${\%}$) & 89(69${\%}$) & $\backslash$ & $\backslash$  \\
        Sex: Male(yes)  & 25(62.5${\%}$) & 59(66.3${\%}$) & ${\chi ^2}(1) = 0.175$ & 0.676 \\
        \makecell[c]{Age in Months\\(Mean±SD)} & 35.51$\pm$9.968 & 33.83$\pm$8.028 & T(127)=0.938 & 0.350 \\     
        \bottomrule
    \end{tabular}
    \label{tab:demographics}
\end{table}















\bibliographystyle{unsrt}

\bibliography{cas_refs}

\begin{thebibliography}{10}

\bibitem{baio2018prevalence}
Jon Baio, Lisa Wiggins, Deborah~L Christensen, Matthew~J Maenner, Julie Daniels, Zachary Warren, Margaret Kurzius-Spencer, Walter Zahorodny, Cordelia~Robinson Rosenberg, Tiffany White, et~al.
\newblock Prevalence of autism spectrum disorder among children aged 8 years—autism and developmental disabilities monitoring network, 11 sites, united states, 2014.
\newblock {\em MMWR Surveillance Summaries}, 67(6):1, 2018.

\bibitem{yuan2022early}
Yuzhuo Yuan and Fang Luo.
\newblock Early screening and diagnosis of autism spectrum disorder assisted by artificial intelligence.
\newblock {\em Advances in Psychological Science}, 30(10):2303--2320, 2022.

\bibitem{wallace2012global}
Simon Wallace, Deborah Fein, Michael Rosanoff, Geraldine Dawson, Saima Hossain, Lynn Brennan, Ariel Como, and Andy Shih.
\newblock A global public health strategy for autism spectrum disorders.
\newblock {\em Autism Research}, 5(3):211--217, 2012.

\bibitem{dawson1998children}
Geraldine Dawson, Andrew~N Meltzoff, Julie Osterling, Julie Rinaldi, and Emily Brown.
\newblock Children with autism fail to orient to naturally occurring social stimuli.
\newblock {\em Journal of autism and developmental disorders}, 28:479--485, 1998.

\bibitem{dawson2004early}
Geraldine Dawson, Karen Toth, Robert Abbott, Julie Osterling, Jeff Munson, Annette Estes, and Jane Liaw.
\newblock Early social attention impairments in autism: social orienting, joint attention, and attention to distress.
\newblock {\em Developmental psychology}, 40(2):271--283, 2004.

\bibitem{werner2000brief}
Emily Werner, Geraldine Dawson, Julie Osterling, and Nuhad Dinno.
\newblock Brief report: Recognition of autism spectrum disorder before one year of age: A retrospective study based on home videotapes.
\newblock {\em Journal of autism and developmental disorders}, 30(2):157, 2000.

\bibitem{constantino2017infant}
John~N Constantino, Stefanie Kennon-McGill, Claire Weichselbaum, Natasha Marrus, Alyzeh Haider, Anne~L Glowinski, Scott Gillespie, Cheryl Klaiman, Ami Klin, and Warren Jones.
\newblock Infant viewing of social scenes is under genetic control and is atypical in autism.
\newblock {\em Nature}, 547(7663):340--344, 2017.

\bibitem{crowell2019parenting}
Judith~A Crowell, Jennifer Keluskar, and Amanda Gorecki.
\newblock Parenting behavior and the development of children with autism spectrum disorder.
\newblock {\em Comprehensive psychiatry}, 90:21--29, 2019.

\bibitem{dunlap2019autism}
Jayne~Jennings Dunlap.
\newblock Autism spectrum disorder screening and early action.
\newblock {\em The Journal for Nurse Practitioners}, 15(7):496--501, 2019.

\bibitem{dilavore1995pre}
Pamela~C DiLavore, Catherine Lord, and Michael Rutter.
\newblock The pre-linguistic autism diagnostic observation schedule.
\newblock {\em Journal of autism and developmental disorders}, 25(4):355--379, 1995.

\bibitem{lord1994autism}
Catherine Lord, Michael Rutter, and Ann Le~Couteur.
\newblock Autism diagnostic interview-revised: a revised version of a diagnostic interview for caregivers of individuals with possible pervasive developmental disorders.
\newblock {\em Journal of autism and developmental disorders}, 24(5):659--685, 1994.

\bibitem{falkmer2013diagnostic}
Torbj{\"o}rn Falkmer, Katie Anderson, Marita Falkmer, and Chiara Horlin.
\newblock Diagnostic procedures in autism spectrum disorders: a systematic literature review.
\newblock {\em European child \& adolescent psychiatry}, 22:329--340, 2013.

\bibitem{randall2018diagnostic}
Melinda Randall, Kristine~J Egberts, Aarti Samtani, Rob~JPM Scholten, Lotty Hooft, Nuala Livingstone, Katy Sterling-Levis, Susan Woolfenden, and Katrina Williams.
\newblock Diagnostic tests for autism spectrum disorder (asd) in preschool children.
\newblock {\em Cochrane Database of Systematic Reviews}, 2018.

\bibitem{liu2020early}
Jingjing Liu, Zhiyong Wang, Kai Xu, Bin Ji, Gongyue Zhang, Yi~Wang, Jingxin Deng, Qiong Xu, Xiu Xu, and Honghai Liu.
\newblock Early screening of autism in toddlers via response-to-instructions protocol.
\newblock {\em IEEE Transactions on Cybernetics}, 52(5):3914--3924, 2022.

\bibitem{wang2019ENIFP}
Zhiyong Wang, Kai Xu, and Honghai Liu.
\newblock Screening early children with autism spectrum disorder via expressing needs with index finger pointing.
\newblock In {\em Proceedings of the 13th International Conference on Distributed Smart Cameras}, pages 1--6, 2019.

\bibitem{marinoiu20183d}
Elisabeta Marinoiu, Mihai Zanfir, Vlad Olaru, and Cristian Sminchisescu.
\newblock 3d human sensing, action and emotion recognition in robot assisted therapy of children with autism.
\newblock In {\em Proceedings of the IEEE conference on computer vision and pattern recognition}, pages 2158--2167, 2018.

\bibitem{murias2018validation}
Michael Murias, Samantha Major, Katherine Davlantis, Lauren Franz, Adrianne Harris, Benjamin Rardin, Maura Sabatos-DeVito, and Geraldine Dawson.
\newblock Validation of eye-tracking measures of social attention as a potential biomarker for autism clinical trials.
\newblock {\em Autism Research}, 11(1):166--174, 2018.

\bibitem{martin2018objective}
Katherine~B Martin, Zakia Hammal, Gang Ren, Jeffrey~F Cohn, Justine Cassell, Mitsunori Ogihara, Jennifer~C Britton, Anibal Gutierrez, and Daniel~S Messinger.
\newblock Objective measurement of head movement differences in children with and without autism spectrum disorder.
\newblock {\em Molecular autism}, 9:1--10, 2018.

\bibitem{negin2021vision}
Farhood Negin, Baris Ozyer, Saeid Agahian, Sibel Kacdioglu, and Gulsah~Tumuklu Ozyer.
\newblock Vision-assisted recognition of stereotype behaviors for early diagnosis of autism spectrum disorders.
\newblock {\em Neurocomputing}, 446:145--155, 2021.

\bibitem{cordeiro2020evaluating}
Lisa Cordeiro, Marcia Braden, Elizabeth Coan, Nanastasia Welnick, Tanea Tanda, and Nicole Tartaglia.
\newblock Evaluating social interactions using the autism screening instrument for education planning-(asiep-3): interaction assessment in children and adults with fragile x syndrome.
\newblock {\em Brain sciences}, 10(4):248, 2020.

\bibitem{stone2000brief}
Wendy~L Stone, Elaine~E Coonrod, and Opal~Y Ousley.
\newblock Brief report: screening tool for autism in two-year-olds (stat): development and preliminary data.
\newblock {\em Journal of autism and developmental disorders}, 30(6):607, 2000.

\bibitem{chen2019attention}
Shi Chen and Qi~Zhao.
\newblock Attention-based autism spectrum disorder screening with privileged modality.
\newblock In {\em 2019 IEEE/CVF International Conference on Computer Vision}, pages 1181--1190, 2019.

\bibitem{li2020classifying}
Jing Li, Yihao Zhong, Junxia Han, Gaoxiang Ouyang, Xiaoli Li, and Honghai Liu.
\newblock Classifying asd children with lstm based on raw videos.
\newblock {\em Neurocomputing}, 390:226--238, 2020.

\bibitem{alvari2021smiling}
Gianpaolo Alvari, Cesare Furlanello, and Paola Venuti.
\newblock Is smiling the key? machine learning analytics detect subtle patterns in micro-expressions of infants with asd.
\newblock {\em Journal of clinical medicine}, 10(8):1776, 2021.

\bibitem{campbell2019computer}
Kathleen Campbell, Kimberly~LH Carpenter, Jordan Hashemi, Steven Espinosa, Samuel Marsan, Jana~Schaich Borg, Zhuoqing Chang, Qiang Qiu, Saritha Vermeer, Elizabeth Adler, et~al.
\newblock Computer vision analysis captures atypical attention in toddlers with autism.
\newblock {\em Autism}, 23(3):619--628, 2019.

\bibitem{qin2021vision}
Haibo Qin, Zhiyong Wang, Jingjing Liu, Qiong Xu, Huiping Li, Xiu Xu, and Honghai Liu.
\newblock Vision-based pointing estimation and evaluation in toddlers for autism screening.
\newblock In {\em Intelligent Robotics and Applications}, pages 177--185. Springer, 2021.

\bibitem{washington2021activity}
Peter Washington, Aaron Kline, Onur~Cezmi Mutlu, Emilie Leblanc, Cathy Hou, Nate Stockham, Kelley Paskov, Brianna Chrisman, and Dennis Wall.
\newblock Activity recognition with moving cameras and few training examples: applications for detection of autism-related headbanging.
\newblock In {\em Extended abstracts of the 2021 CHI conference on human factors in computing systems}, pages 1--7, 2021.

\bibitem{wang2019screening}
Zhiyong Wang, Jingjing Liu, Keshi He, Qiong Xu, Xiu Xu, and Honghai Liu.
\newblock Screening early children with autism spectrum disorder via response-to-name protocol.
\newblock {\em IEEE Transactions on Industrial Informatics}, 17(1):587--595, 2021.

\bibitem{zunino2018video}
Andrea Zunino, Pietro Morerio, Andrea Cavallo, Caterina Ansuini, Jessica Podda, Francesca Battaglia, Edvige Veneselli, Cristina Becchio, and Vittorio Murino.
\newblock Video gesture analysis for autism spectrum disorder detection.
\newblock In {\em 2018 24th international conference on pattern recognition (ICPR)}, pages 3421--3426. IEEE, 2018.

\bibitem{hong2020toward}
Seok-Jun Hong, Joshua~T Vogelstein, Alessandro Gozzi, Boris~C Bernhardt, BT~Thomas Yeo, Michael~P Milham, and Adriana Di~Martino.
\newblock Toward neurosubtypes in autism.
\newblock {\em Biological Psychiatry}, 88(1):111--128, 2020.

\bibitem{min2009detection}
Cheol-Hong Min, Youngchun Kim, Ahmed Tewfik, and Anne Kelly.
\newblock Detection of self-stimulatory behaviors of children with autism using wearable and environmental sensors.
\newblock {\em Journal of Medical Devices}, 3(2):027506, 2009.

\bibitem{westeyn2005recognizing}
Tracy Westeyn, Kristin Vadas, Xuehai Bian, Thad Starner, and Gregory~D Abowd.
\newblock Recognizing mimicked autistic self-stimulatory behaviors using hmms.
\newblock In {\em Ninth IEEE International Symposium on Wearable Computers (ISWC'05)}, pages 164--167. IEEE, 2005.

\bibitem{al2020generating}
Ahmed~A Al-Jubouri, Israa~Hadi Ali, and Yasen Rajihy.
\newblock Generating 3d dataset of gait and full body movement of children with autism spectrum disorders collected by kinect v2 camera.
\newblock {\em Compusoft}, 9(8):3791--3797, 2020.

\bibitem{ali2022video}
Abid Ali, Farhood~F Negin, Francois~F Bremond, and Susanne Th{\"u}mmler.
\newblock Video-based behavior understanding of children for objective diagnosis of autism.
\newblock In {\em Proceedings of the 17th International Joint Conference on Computer Vision, Imaging and Computer Graphics Theory and Applications}, pages 475--484, 2022.

\bibitem{pandeyGuidedWeakSupervision2020a}
Prashant Pandey, AP~Prathosh, Manu Kohli, and Josh Pritchard.
\newblock Guided weak supervision for action recognition with scarce data to assess skills of children with autism.
\newblock In {\em Proceedings of the AAAI Conference on Artificial Intelligence}, volume~34, pages 463--470, 2020.

\bibitem{tianVideoBasedEarlyASD2019a}
Yuan Tian, Xiongkuo Min, Guangtao Zhai, and Zhiyong Gao.
\newblock Video-based early asd detection via temporal pyramid networks.
\newblock In {\em 2019 IEEE International Conference on Multimedia and Expo (ICME)}, pages 272--277. IEEE, 2019.

\bibitem{xu2024autism}
Yongjie Xu, Zengjie Yu, Yisheng Li, Yuehan Liu, Ye~Li, and Yishan Wang.
\newblock Autism spectrum disorder diagnosis with eeg signals using time series maps of brain functional connectivity and a combined cnn--lstm model.
\newblock {\em Computer Methods and Programs in Biomedicine}, 250:108196, 2024.

\bibitem{wang2024residual}
Yibin Wang, Haixia Long, Tao Bo, and Jianwei Zheng.
\newblock Residual graph transformer for autism spectrum disorder prediction.
\newblock {\em Computer Methods and Programs in Biomedicine}, 247:108065, 2024.

\bibitem{rajagopalan2014detecting}
Shyam~Sundar Rajagopalan and Roland Goecke.
\newblock Detecting self-stimulatory behaviours for autism diagnosis.
\newblock In {\em 2014 IEEE International Conference on Image Processing (ICIP)}, pages 1470--1474. IEEE, 2014.

\bibitem{schopler2010childhood}
Eric Schopler, Robert~Jay Reichler, and Barbara~Rochen Renner.
\newblock {\em The childhood autism rating scale (CARS)}.
\newblock Western Psychological Services Los Angeles, CA, 2010.

\bibitem{world2001world}
World~Medical Association et~al.
\newblock World medical association declaration of helsinki. ethical principles for medical research involving human subjects.
\newblock {\em JAMA}, 310(20):2191--2194, 11 2013.

\bibitem{ren2015faster}
Shaoqing Ren, Kaiming He, Ross Girshick, and Jian Sun.
\newblock Faster r-cnn: Towards real-time object detection with region proposal networks.
\newblock {\em IEEE Transactions on Pattern Analysis and Machine Intelligence}, 39(6):1137--1149, 2017.

\bibitem{sun2019deep}
Ke~Sun, Bin Xiao, Dong Liu, and Jingdong Wang.
\newblock Deep high-resolution representation learning for human pose estimation.
\newblock In {\em 2019 IEEE/CVF Conference on Computer Vision and Pattern Recognition (CVPR)}, pages 5686--5696. IEEE, 2019.

\bibitem{yan2018spatial}
Sijie Yan, Yuanjun Xiong, and Dahua Lin.
\newblock Spatial temporal graph convolutional networks for skeleton-based action recognition.
\newblock In {\em Proceedings of the AAAI conference on artificial intelligence}, volume~32, pages 7444--7452, 2018.

\bibitem{beck2024xlstm}
Maximilian Beck, Korbinian P{\"o}ppel, Markus Spanring, Andreas Auer, Oleksandra Prudnikova, Michael Kopp, G{\"u}nter Klambauer, Johannes Brandstetter, and Sepp Hochreiter.
\newblock xlstm: Extended long short-term memory.
\newblock {\em arXiv preprint arXiv:2405.04517}, 2024.

\bibitem{kipf2016semi}
Thomas~N Kipf and Max Welling.
\newblock Semi-supervised classification with graph convolutional networks.
\newblock {\em arXiv preprint arXiv:1609.02907}, 2016.

\bibitem{wang2019adaptively}
Guangrun Wang, Keze Wang, and Liang Lin.
\newblock Adaptively connected neural networks.
\newblock In {\em 2019 IEEE/CVF Conference on Computer Vision and Pattern Recognition (CVPR)}, pages 1781--1790, 2019.

\bibitem{hochreiter1997long}
Sepp Hochreiter and J{\"u}rgen Schmidhuber.
\newblock Long short-term memory.
\newblock {\em Neural computation}, 9(8):1735--1780, 1997.

\bibitem{alkin2024vision}
Benedikt Alkin, Maximilian Beck, Korbinian P{\"o}ppel, Sepp Hochreiter, and Johannes Brandstetter.
\newblock Vision-lstm: xlstm as generic vision backbone.
\newblock {\em arXiv preprint arXiv:2406.04303}, 2024.

\bibitem{nie2017monocular}
Bruce~Xiaohan Nie, Ping Wei, and Song-Chun Zhu.
\newblock Monocular 3d human pose estimation by predicting depth on joints.
\newblock In {\em 2017 IEEE International Conference on Computer Vision (ICCV)}, pages 3467--3475. IEEE, 2017.

\bibitem{duan2022revisiting}
Haodong Duan, Yue Zhao, Kai Chen, Dahua Lin, and Bo~Dai.
\newblock Revisiting skeleton-based action recognition.
\newblock In {\em 2022 IEEE/CVF Conference on Computer Vision and Pattern Recognition}, pages 2969--2978, 2022.

\bibitem{chen2023self}
Xuanchi Chen, Xiangwei Zheng, Kai Sun, Weilong Liu, and Yuang Zhang.
\newblock Self-supervised vision transformer-based few-shot learning for facial expression recognition.
\newblock {\em Information Sciences}, 634:206--226, 2023.

\bibitem{yang2023invertible}
Yang Yang, Yiyang Huang, Ming Shi, Kejiang Chen, and Weiming Zhang.
\newblock Invertible mask network for face privacy preservation.
\newblock {\em Information Sciences}, 629:566--579, 2023.

\bibitem{kim2024self}
Sungjune Kim, Seongjun Yun, Jongwuk Lee, Gyusam Chang, Wonseok Roh, Dae-Neung Sohn, Jung-Tae Lee, Hogun Park, and Sangpil Kim.
\newblock Self-supervised multimodal graph convolutional network for collaborative filtering.
\newblock {\em Information Sciences}, 653:119760, 2024.

\bibitem{ke2024two}
Xiao Ke, Huangbiao Xu, Xiaofeng Lin, and Wenzhong Guo.
\newblock Two-path target-aware contrastive regression for action quality assessment.
\newblock {\em Information Sciences}, 664:120347, 2024.

\end{thebibliography}



\end{document}